\documentclass[runningheads]{llncs}
\usepackage[T1]{fontenc}
\usepackage{graphicx}
\usepackage{hyperref}
\begin{document}
\title{ Comparative of Genetic Fuzzy regression techniques for aero-acoustic phenomenons }
%
%
\author{Hugo Henry\inst{1}\orcidID{0009-0001-2530-3521}\and 
Dr. Kelly Cohen\inst{4}\orcidID{0000-0002-8655-1465}}
\authorrunning{H. Henry and K. Cohen}
%
\institute{University of Cincinnati, Cincinnati OH 45221, USA\\
\email{henryho@mail.uc.edu}\\
\email{cohenky@ucmail.uc.edu}}
\maketitle              
%


\begin{abstract}
This study investigates the application of Genetic Fuzzy Systems (GFS) to model the self-noise generated by airfoils, a key issue in aeroacoustics with significant implications for aerospace, automotive, and drone applications. Using the publicly available "Airfoil Self Noise" dataset, various fuzzy regression strategies are explored and compared. The paper evaluates a brute-force Takagi-Sugeno-Kang (TSK) fuzzy system with high rule density, a cascading Genetic Fuzzy Tree (GFT) architecture, and a novel clustered approach based on Fuzzy C-Means (FCM) to reduce the model's complexity.
This highlights the viability of clustering-assisted fuzzy inference as an effective regression tool for complex aero-acoustic phenomena.

\end{abstract}

\section{Introduction}
Airfoils have a lot of application, especially in aerospace but in a more general sense in aerodynamics, including aerospace, automotive and naval industry. Even though they have a lot of use and a lot of advantages, there is also drawbacks to the extensive use of airfoils. One of the most notable is vibrations, an airfoil possesses different vibration modes depending on the geometry and the materials of the airfoil.
Vibration creates two unwanted phenomena, constraints and resonance on the structure potentially leading to problems in structural integrity and noise for the people near the airfoil.

The noise in particular is something that have to be studied in order for the foil to be used in crowded area, notably helicopters, drones or formula cars.

\section{Survey Methodology}

This study will use the dataset "Airfoil Self Noise"\cite{dataset} found on the UCI Database.

\section{analysis of the dataset}
\label{subsec:analysis}
The dataset used for this regression is composed of 1503 points along 6 features, 5 inputs and 1 output:
\begin{itemize}
    \item Frequency $\in [50-20000]$ Hz
    \item angle of attack $\in [0-22.2]$° (discrete set)
    \item chord length $\in [0.254-0.3048]$ (discrete set)
    \item free stream Velocity $\in [31.7-71.3]$ (discrete set)
    \item suction side displacement thickness $\in [0.00040068-0.0584]$
    \item Noise in dB $\in [103.38-140.987]$
\end{itemize}

\begin{figure}[h]
    \centering
    \includegraphics[width=1\linewidth]{ 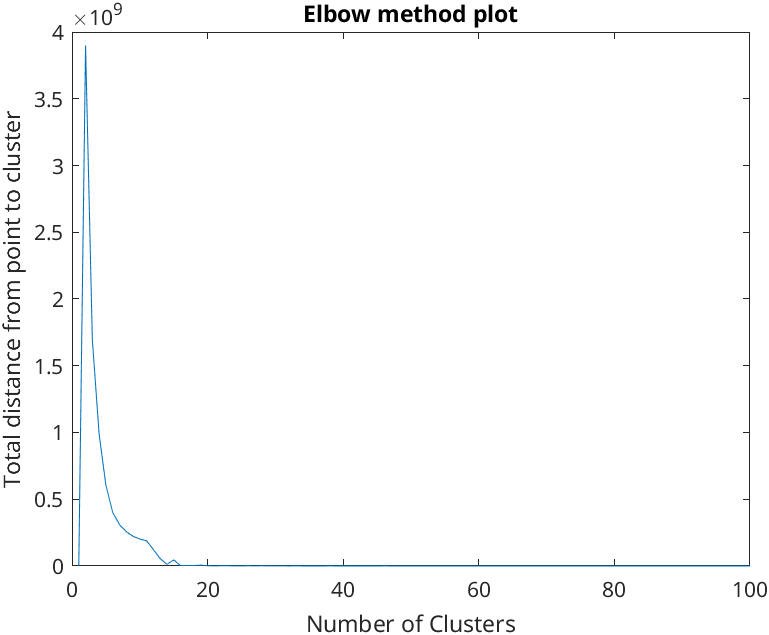}
    \caption{Clustering of the dataset}
    \label{fig:clustering}
\end{figure}

For further insight, the dataset underwent a clustering methodology, here we used the Fuzzy C-means (FCM) alongside the elbow method to determine the number of clusters present in this dataset. Referring figure\ref{fig:clustering}.

Based on this method, we chose to use 15 clusters in the following regressions in part \ref{subsec:clustered}.

\section{Fuzzy regressions}
\subsection{Brute force system}
This dataset containing 5 inputs create a huge space of rules possibles for the system and approaches the limits for a simple crossover based Genetic Algorithm (GA).

Indeed, the system used here is composed of 5 membership functions per input and as such a grand total of 3125 rules to be encoded in the chromosome.

Using a TSK (Takagi Sugeno Kang) 1\textsuperscript{st} order, and triangular membership functions, we attain a total of 18825 parameters as shown in equation \ref{eq:number of param}

\begin{equation}
    number\textunderscore{of}\textunderscore{parameters}=3125*6+5*5*3=18825
    \label{eq:number of param}
\end{equation}

This huge chromosome highly constraints the size of the population and the number of generation in order to keep a reasonable training time. For the rest of this section, the population size will be 50 and 100 generation of training.
\begin{figure}[h]
    \centering
    \begin{minipage}{.49\textwidth}
    \centering
    \includegraphics[width =1\textwidth]{ 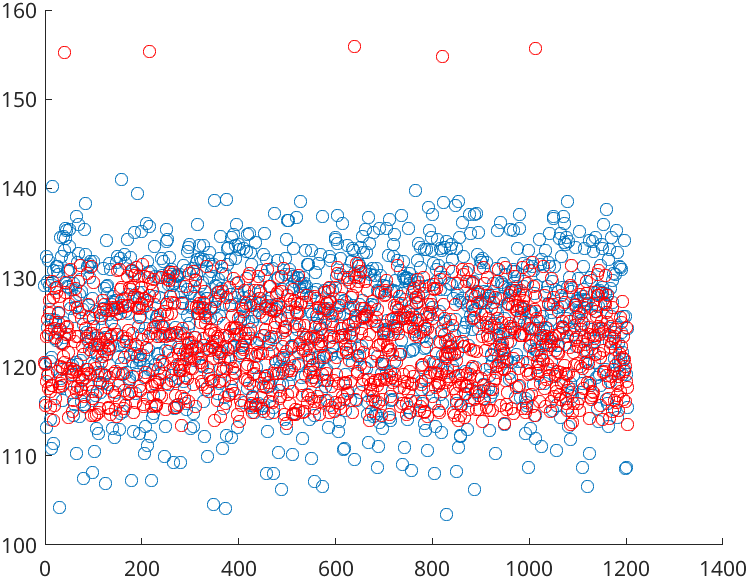}
    \caption{Training set (blue) and prediction (red)}
    \label{fig:bruteforcetrain}
    \end{minipage}
    \begin{minipage}{.49\textwidth}
    \centering
    \includegraphics[width = 1\textwidth]{ 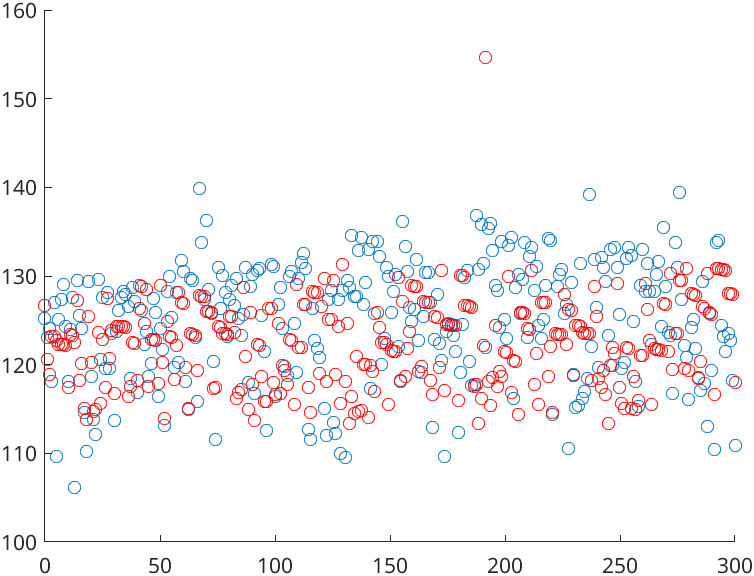}
    \caption{Testing set (blue) and prediction (red)}
    \label{fig:brutforcetest}
    \end{minipage}
    
\end{figure}

\begin{figure}[h]
    \centering
    \includegraphics[width=0.55\linewidth]{ 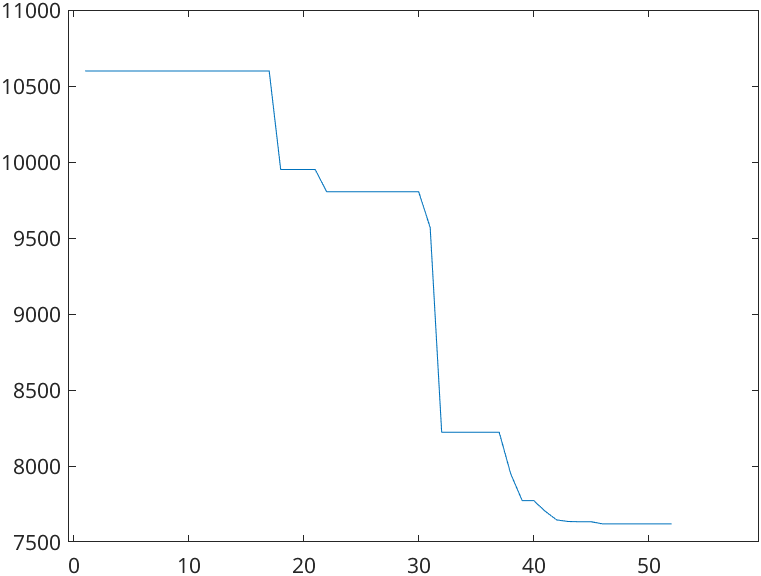}
    \caption{Fitness evolution}
    \label{fig:bruteforcefit}
\end{figure}

We can see that the fitting seems quite adequate for the dataset despite the pretty low number of generations and and population size. indeed we can confidently state that the GA didn't finish optimizing by the few stray training prediction and the mere 8 fitness improvement over the duration of training.
\ref{fig:bruteforcetrain},\ref{fig:brutforcetest} and \ref{fig:bruteforcefit}.

\label{subsec:brutforce}
\newpage

\subsection{Cascading systems}
\subsubsection{TSK 0 order}
\begin{figure}[h]
    \centering
    \includegraphics[width=1\linewidth]{ 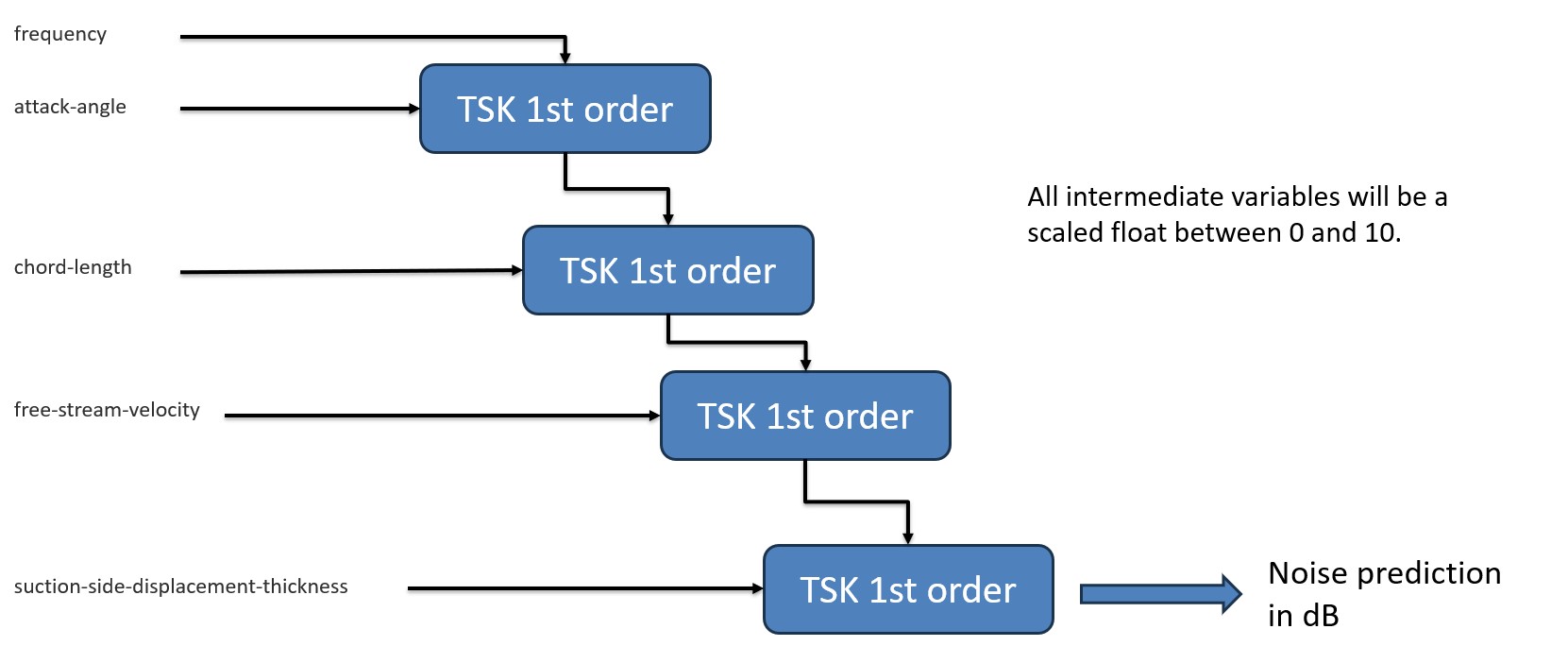}
    \caption{Clustering of the dataset}
    \label{fig:GFT}
\end{figure}

As shown previously, it is crucial to reduce the number of rules in order to reduce the number of parameters and improve training efficiency.
As such, we tried a Genetic Fuzzy Tree (GFT) architecture instead of the brute force to reduce the number of rules to train. This GFT will use the architecture presented in figure \ref{fig:GFT}.

In order to compare, this inference method will use 3 membership function per input and then 5 membership functions per input. allowing comparison with part \ref{subsec:brutforce}.

\begin{figure}[h]
    \centering
    \begin{minipage}{.49\textwidth}
    \centering
    \includegraphics[width =1\textwidth]{ 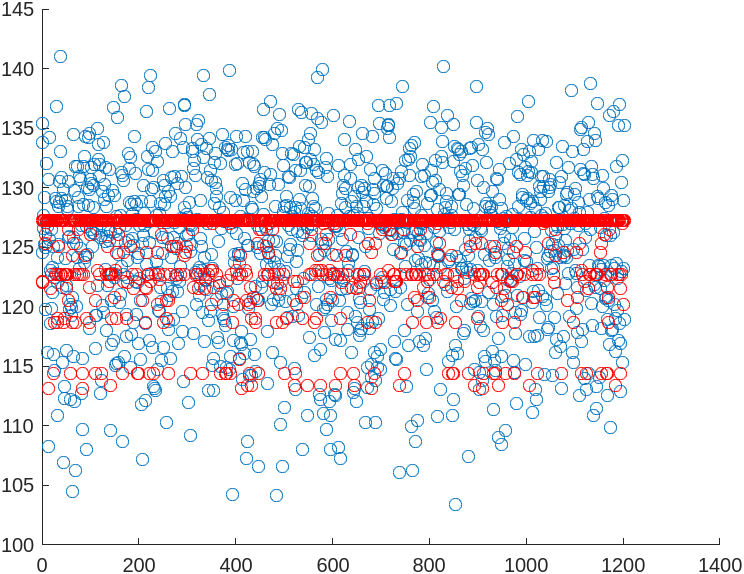}
    \caption{Training set (blue) and prediction (red)}
    \label{fig:3mf0train}
    \end{minipage}
    \begin{minipage}{.49\textwidth}
    \centering
    \includegraphics[width = 1\textwidth]{ 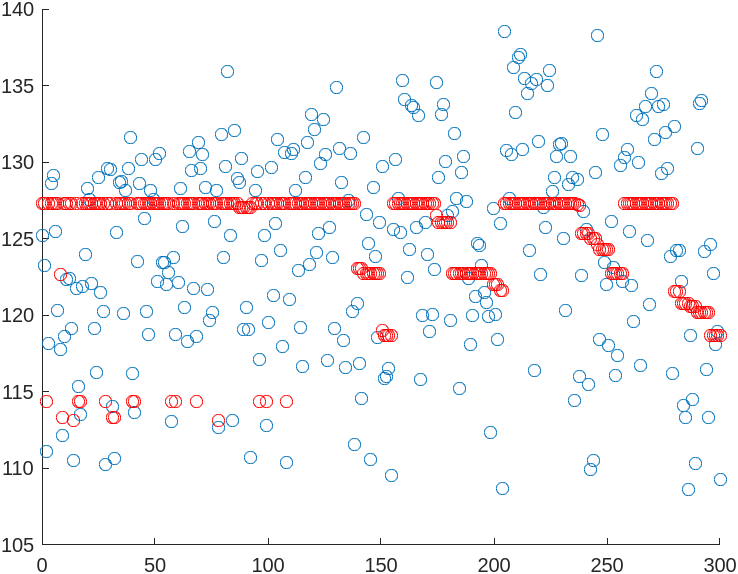}
    \caption{Testing set (blue) and prediction (red)}
    \label{fig:3mf0test}
    \end{minipage}
    
\end{figure}

\begin{figure}[h]
    \centering
    \includegraphics[width=0.55\linewidth]{ 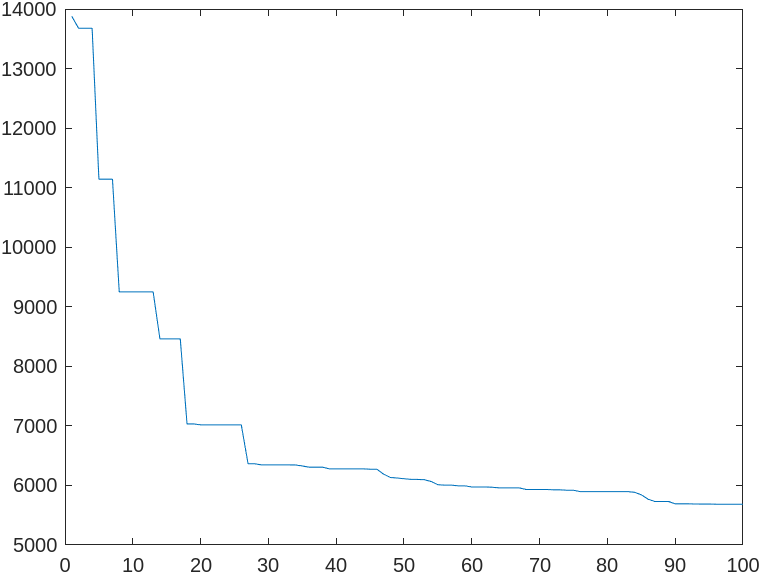}
    \caption{Fitness evolution}
    \label{fig:3mf0fit}
\end{figure}

We can see that the prediction is very longitudinal with very few change which would mean that the different rules do not really intersect with each other and implies a very granular dataset. as such we observe more of a trend prediction than a true prediction. Referring figures \ref{fig:3mf0train} and \ref{fig:3mf0test}

Although this prediction do not make physical sense due to the nature of the output being in Decibels and therefore a linear trend line over a logarithmic value dot make sense we can see the work of the GFT trying to find the best distribution. Referring figure \ref{fig:3mf0fit}.\\

The GFT losing explainability because of the intermediate nodes it is very difficult to understand and therefore debugg such a result outside of trying to add more membership functions and rules to solve this non-overlapping situation.

So did we with 5 membership functions instead of 3.

\begin{figure}[h]
    \centering
    \begin{minipage}{.49\textwidth}
    \centering
    \includegraphics[width =1\textwidth]{ 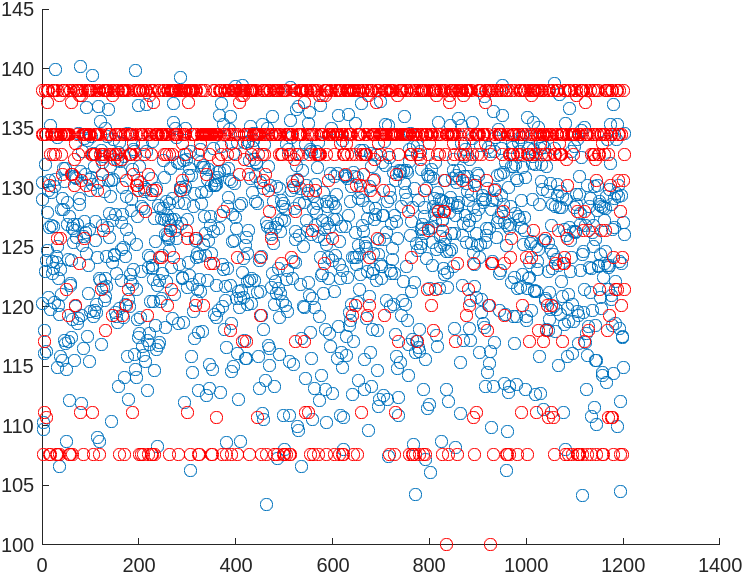}
    \caption{Training set (blue) and prediction (red)}
    \label{fig:5mf0train}
    \end{minipage}
    \begin{minipage}{.49\textwidth}
    \centering
    \includegraphics[width = 1\textwidth]{ 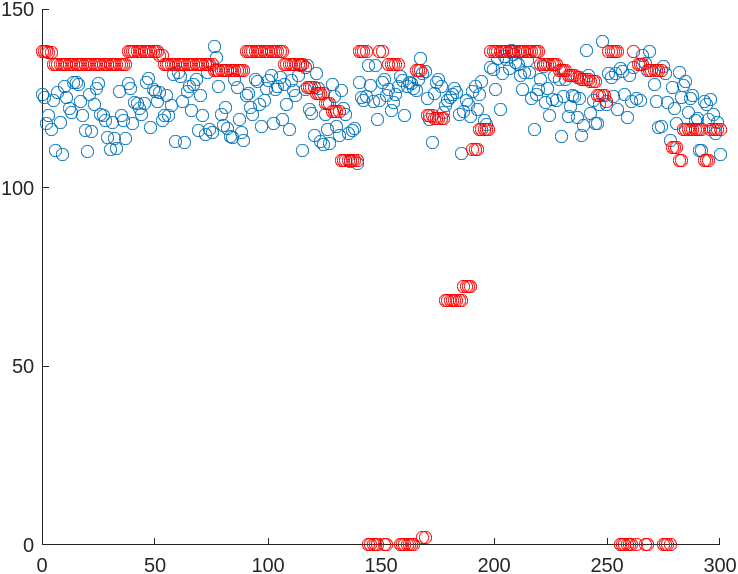}
    \caption{Testing set (blue) and prediction (red)}
    \label{fig:5mf0test}
    \end{minipage}
    
\end{figure}

\begin{figure}[h]
    \centering
    \includegraphics[width=0.55\linewidth]{ 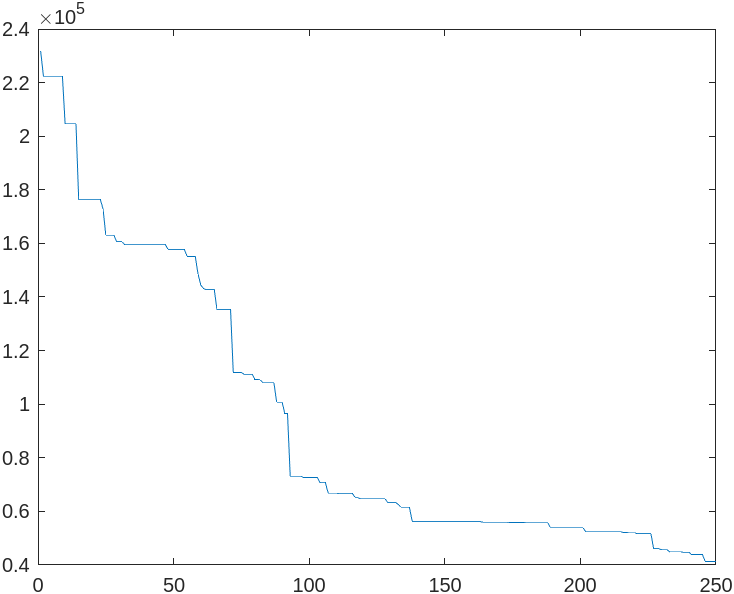}
    \caption{Fitness evolution}
    \label{fig:5mf0fit}
\end{figure}

This time we see a bit more overlapping but the prediction is still very much longitudinal and on the test set, there is even null values which shows us that this model using the 0 order TSK cannot be trusted for this dataset. referring figures \ref{fig:5mf0train}, \ref{fig:5mf0test} and \ref{fig:5mf0fit}.

\subsubsection{TSK 1\textsuperscript{st} order}
In order to escape this longitudinal trending, we then applied a 1\textsuperscript{st} order to get slopes and try to avoid this behavior.

In the same spirit than for a first order, we ran for 3 then 5 membership functions.

\begin{figure}[h]
    \centering
    \begin{minipage}{.49\textwidth}
    \centering
    \includegraphics[width =1\textwidth]{ 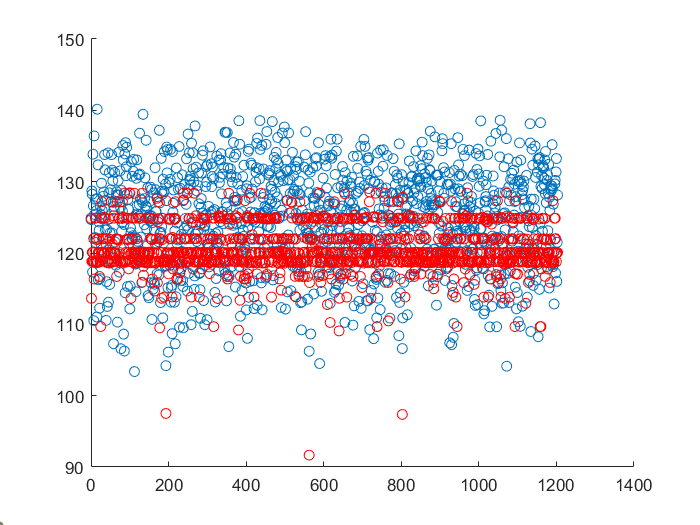}
    \caption{Training set (blue) and prediction (red)}
    \label{fig:3mf1train}
    \end{minipage}
    \begin{minipage}{.49\textwidth}
    \centering
    \includegraphics[width = 1\textwidth]{ 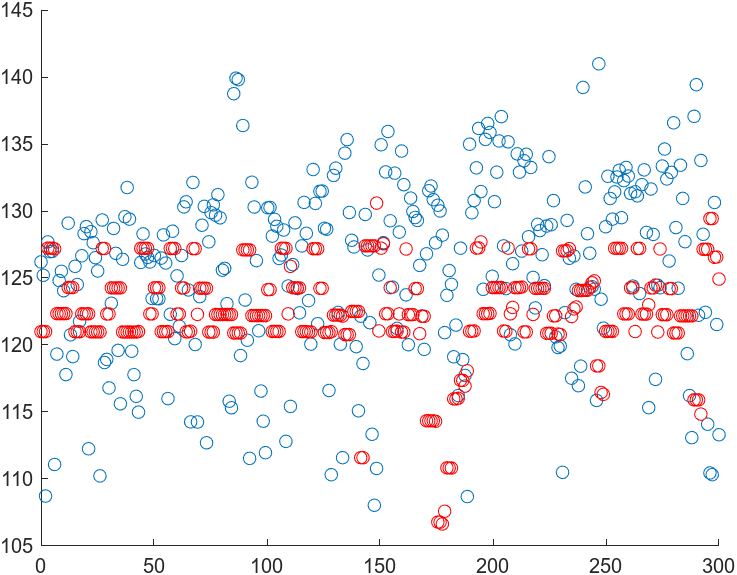}
    \caption{Testing set (blue) and prediction (red)}
    \label{fig:3mf1test}
    \end{minipage}
    
\end{figure}

\begin{figure}[h]
    \centering
    \includegraphics[width=0.55\linewidth]{ 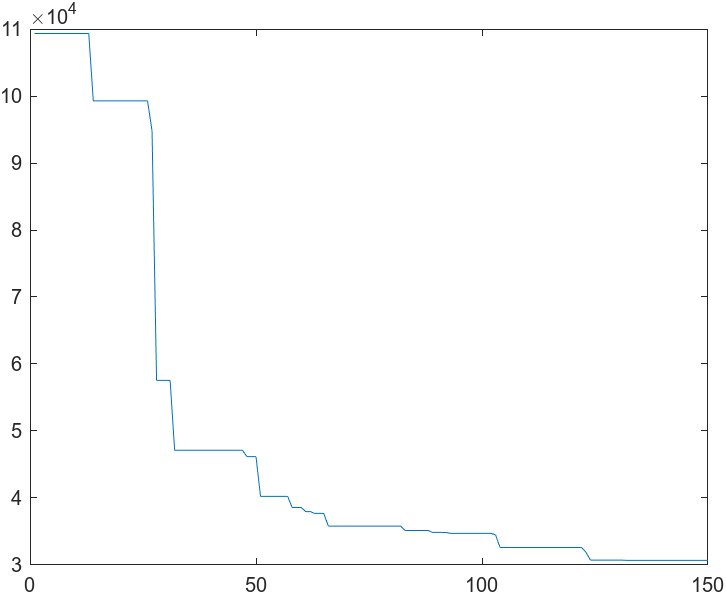}
    \caption{Fitness evolution}
    \label{fig:3mf1fit}
\end{figure}
\newpage

We can see more coverage of the different behaviors in terms of slopes but the behavior seems to continue being blocked by the granularity of the dataset.
Referring figures \ref{fig:3mf1train}, \ref{fig:3mf1test} and \ref{fig:3mf1fit}.

\begin{figure}[h]
    \centering
    \begin{minipage}{.49\textwidth}
    \centering
    \includegraphics[width =1\textwidth]{ 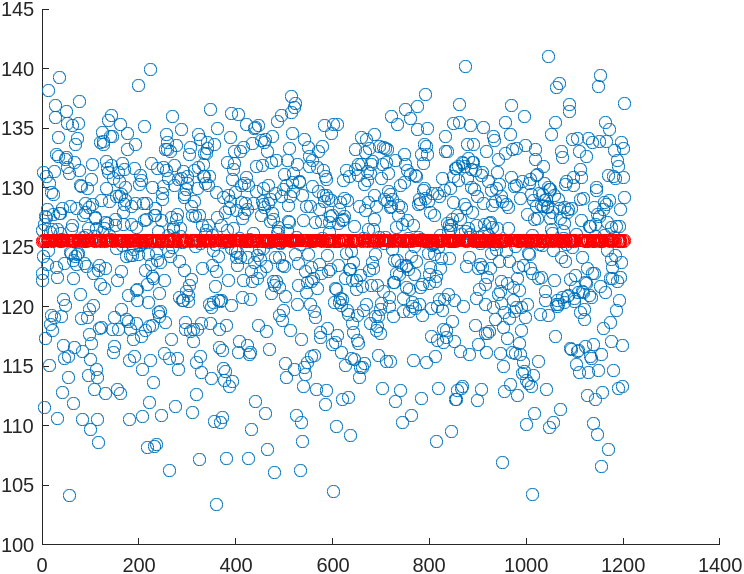}
    \caption{Training set (blue) and prediction (red)}
    \label{fig:5mf1train}
    \end{minipage}
    \begin{minipage}{.49\textwidth}
    \centering
    \includegraphics[width = 1\textwidth]{ 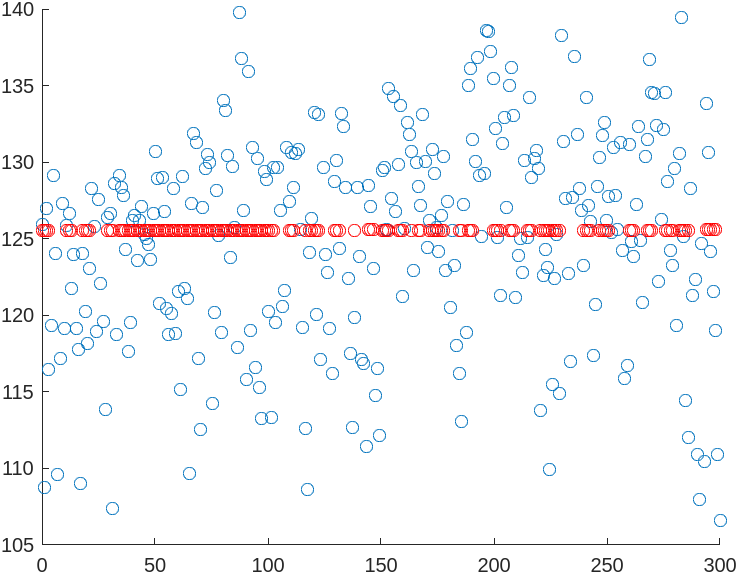}
    \caption{Testing set (blue) and prediction (red)}
    \label{fig:5mf1test}
    \end{minipage}
    
\end{figure}

\begin{figure}[h]
    \centering
    \includegraphics[width=0.55\linewidth]{ 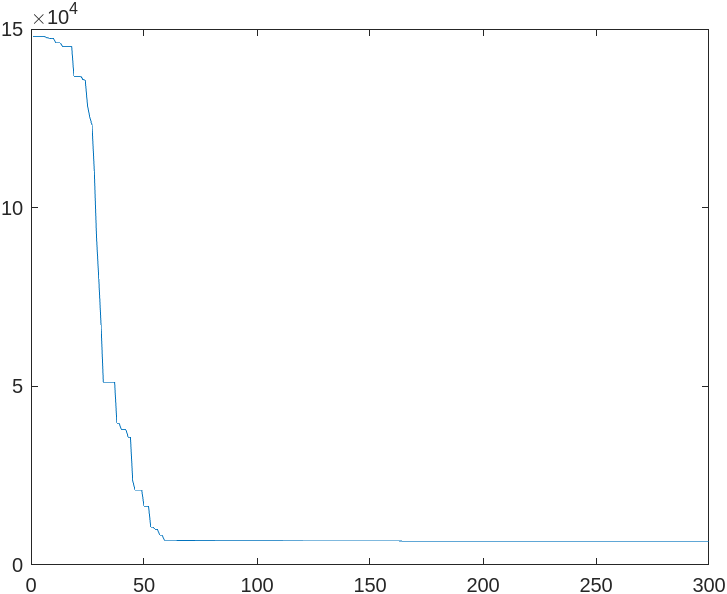}
    \caption{Fitness evolution}
    \label{fig:5mf1fit}
\end{figure}
\newpage

With those 5 membership functions the Genetic algorithm just breaks down completely and goes back to trend lining even with modifications on the fitness function and changing structure in the chromosomes, either it creates a trend line or it goes down to a null vector. Referring figures \ref{fig:5mf1train}, \ref{fig:5mf1test} and \ref{fig:5mf1fit}.

we can conclude this part upon the fact that the cascading system wasn't cut out for this specific dataset and the existence of the intermediate nodes made it very difficult to debugg even with changing the fitness or architecture.

\subsection{Clustered systems}
\label{subsec:clustered}

\begin{figure}[h]
    \centering
    \begin{minipage}{.49\textwidth}
    \centering
    \includegraphics[width =1\textwidth]{ 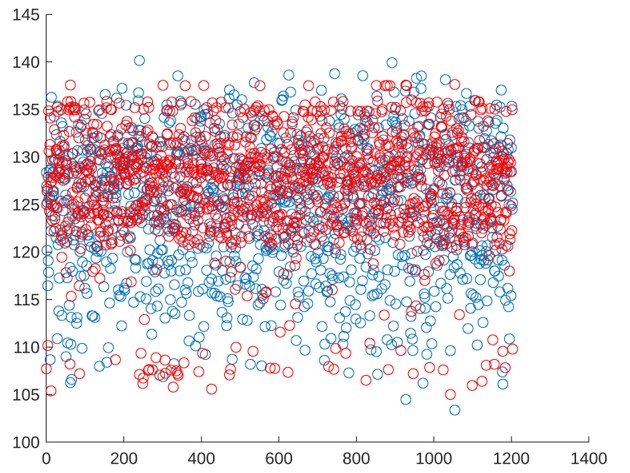}
    \caption{Training set (blue) and prediction (red)}
    \label{fig:cluster1train}
    \end{minipage}
    \begin{minipage}{.49\textwidth}
    \centering
    \includegraphics[width = 1\textwidth]{ 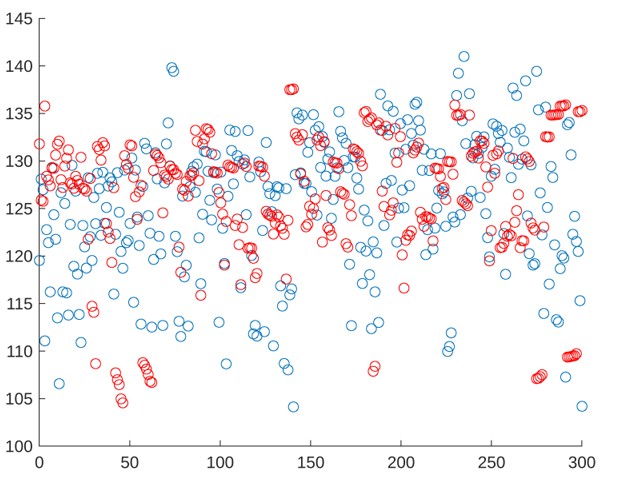}
    \caption{Testing set (blue) and prediction (red)}
    \label{fig:cluster1test}
    \end{minipage}
    
\end{figure}

\begin{figure}[h]
    \centering
    \includegraphics[width=0.55\linewidth]{ 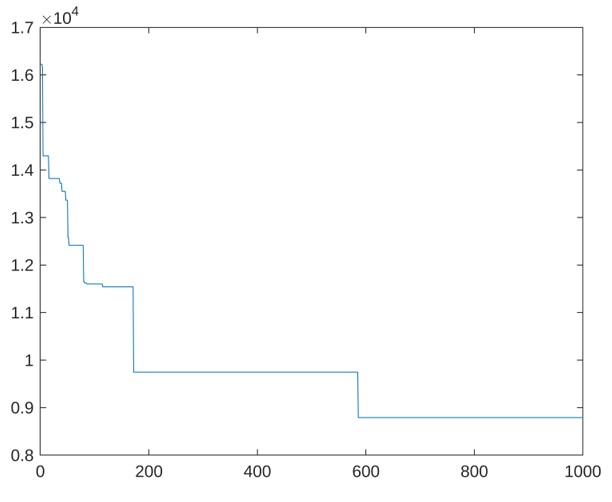}
    \caption{Fitness evolution}
    \label{fig:cluster1fit}
\end{figure}

In order to find a solution to this regression that can be a good approximation like the brute force technique while having a good computing efficiency like the cascading system, we went back to the the analysis done in part \ref{subsec:analysis} and used the clustering to reduce the number of parameters based on the clustering.

for that 2 different inference method have been tried.
first, we created a gaussian membership function centered around the center of each cluster and used them as input, this yielded comfortable results but with a significantly bigger absolute error across the dataset, it made more sense physically but not computationally. Referring to figures \ref{fig:cluster1train}, \ref{fig:cluster1test} and \ref{fig:cluster1fit}.

Then in order to improve this fitness we got rid of the gaussian membership function and took full advantage of the fuzzy clustering technique by recomputing the membership value of the point to the different centers therefore giving the degree of activation of the corresponding rule.

This gave a very good approximation that can be scaled in terms of population size and generations, indeed, the number of parameters being reduced from the 18825 to a mere 90.
\begin{figure}[h]
    \centering
    \begin{minipage}{.49\textwidth}
    \centering
    \includegraphics[width =1\textwidth]{ 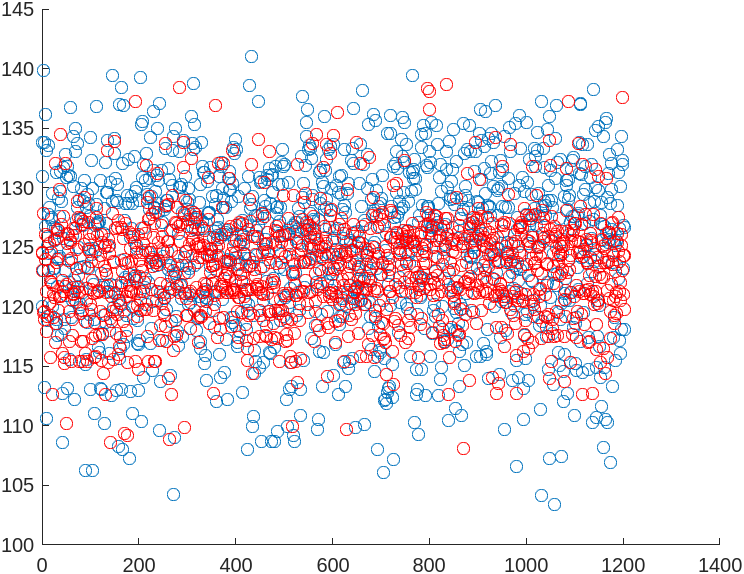}
    \caption{Training set (blue) and prediction (red)}
    \label{fig:5mf1train}
    \end{minipage}
    \begin{minipage}{.49\textwidth}
    \centering
    \includegraphics[width = 1\textwidth]{ 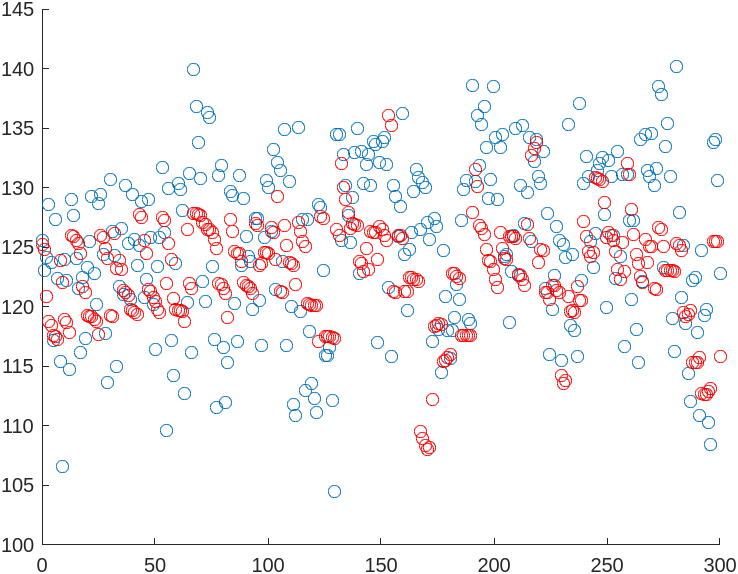}
    \caption{Testing set (blue) and prediction (red)}
    \label{fig:5mf1test}
    \end{minipage}
    
\end{figure}

\begin{figure}[h]
    \centering
    \includegraphics[width=0.55\linewidth]{ 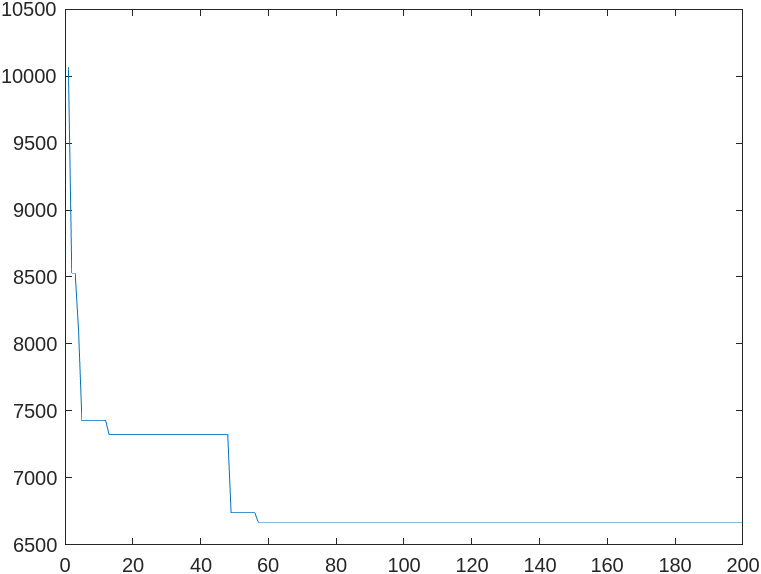}
    \caption{Fitness evolution}
    \label{fig:5mf1fit}
\end{figure}

\newpage
\section{Comparison and analysis of the results}
We can clearly see that in this dataset case, the high granularity of the a few inputs lead to a collapse of the cascading systems and the intermediate nodes make it near impossible to fix due to their virtual nature and of the lack of physical representation. In this particular case, the effect of increasing the order of the TSK inference or the number of membership functions did not change much on the output.\\

For the non-cascading systems, the brute-force method has good precision, but the problem of scalability is evident and therefore not a good solution compared to the clustered system. The clustered approximation will still be worse that the brute-forcing in terms of precision if the time is given to is but the training time is incomparable and for added precision we can increase the number of clusters while keeping a number of parameters very low compared to the brute-force method.

\section{Conclusion}
Through this project, we showed the problem that can be encountered with cascading system, being that if it doesn't work at first, it will be extremely difficult to fix especially toward granularity in the dataset, we also demonstrated first hand the problem of scalability of basic Genetic Fuzzy Systems while still showing that the prediction keep good results despite the granularity. \\

Lastly we showed that clustering have a huge impact in the prediction allowing for the precision of the brute-force while faster than a cascading system and more robust toward the type of data. Moreover, the Fuzzy clustering allow to be less impacted by outliers but still taking them into accounts with the membership values.

Results demonstrate that while brute-force models offer accurate predictions at high computational costs, cascading systems suffer from limited explainability and convergence issues. In contrast, the clustered GFS significantly reduces the number of parameters while maintaining good predictive accuracy, striking a balance between computational efficiency and physical relevance.

Further work would include the use of better clustering methods such as VAT and IVAT as well as comparing other optimization methods such as gradient based optimization or swarm theory.

%
%
%

\end{document}